# Novel scorpion detection system combining computer vision and fluorescence

**Francisco L. Giambelluca[1], Jorge Osio[1,2], Luis A. Giambelluca[3,4], Marcelo A. Cappelletti[1,2,3]**

[1] Grupo de Control Aplicado (GCA), Área CeTAD, Instituto LEICI (CONICET-UNLP). Calle 116 y 48, 2° piso, La Plata (1900), Argentina.

[2] Programa TICAPPS, Universidad Nacional Arturo Jauretche, Av. Calchaquí 6200, Florencio Varela (1888), Argentina

[3] Consejo Nacional de Investigaciones Científicas y Técnicas (CONICET), Argentina

[4] CEPAVE (CONICET-UNLP-CCT La Plata, Asoc. CICBA). Boulevard 120 e/61 y 64, La Plata (1900), Argentina.

E-mail: francisco.giambelluca@ing.unlp.edu.ar; jorge.osio@ing.unlp.edu.ar; giambelluca@cepave.edu.ar; marcelo.cappelletti@ing.unlp.edu.ar

**Abstract**

In this work, a fully automatic and real-time system for the detection of scorpions was developed using computer vision and deep learning techniques. This system is based on the implementation of a double validation process using the shape features and the fluorescent characteristics of scorpions when exposed to ultraviolet (UV) light. The Haar Cascade Classifier (HCC) and YOLO (You Only Look Once) models have been used and compared as the first mechanism for the scorpion shape detection. The detection of the fluorescence emitted by the scorpions under UV light has been used as a second detection mechanism in order to increase the accuracy and precision of the system. The results obtained show that the system can accurately and reliably detect the presence of scorpions. In addition, values obtained of recall of 100% is essential with the purpose of providing a health security tool. Although the developed system can only be used at night or in dark environment, where the fluorescence emitted by the scorpions can be visualized, the nocturnal activity of scorpions justifies the incorporation of this second validation mechanism.

**1. Introduction**

Although scorpion stings mostly occur due to accidental encounters with these arachnids, scorpions have been the object of study for many years, due to their high danger [1]. For this reason, the development of efficient techniques for their detection and identification becomes substantially important.

Scorpions are nocturnal animals with negative phototropism. They can be found sheltered during the day, in natural environments, under rocks or inside holes, although some species can also be domiciliary, which remain hidden in the daytime and roam at night.

In general, the methods used for scorpion detection are dangerous, time-consuming, and invasive, such as rock rolling, burrowing detection, peeling loose back of tree and pitfall trap [2,3]. Other more efficient methods than those mentioned previously, consider biological characteristics of scorpions, such as vibration signals or the fluorescent property. On the one hand, a scorpion detection system using vibration frequency detection technique has been developed in [4], since the scorpion like other arthropods use substrate vibration signals to recognize and locate mates and preys [5,6]. On the other hand, the cuticle of scorpions emit a cyan fluorescence (with wavelength between 440 and 490 nm) when are exposed to ultraviolet (UV) light [7,8]. This phenomenon, which was discovered almost simultaneously by the Italian zoologist M. Pavan and the South African zoologist R.F. Lawrence in 1954, has revolutionized the study of the biology and ecology of scorpions, thanks to the fact that it was possible to detect them at night, in a relatively non-invasive way, using a portable UV light [9-11]. Fig. 1 shows two images of the scorpion *Tityus trivittatus* under natural light (left) and under UV light (right), where the fluorescent property can be observed.

The fluorescence in scorpions is due to the existence of two chemical compounds in the cuticle, such as the alkaloid β-carboline and 7-hydroxy-4-methylcoumarin. These two compounds are found in the hyaline exocuticle, a region of the cuticle that is approximately 4 microns thick in scorpions. The cuticle is a kind of skin that protects arthropods and allows them to maintain their shape.

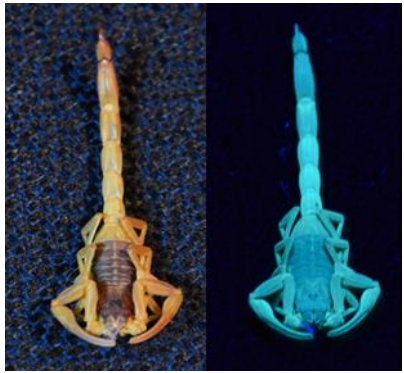

**Figure 1.** Pictures of a *Tityus trivittatus* scorpion under natural light (left) and UV light (right).

The fluorescence intensity varies widely among different species of scorpion and increases with age and with the hardness of the cuticle [12-14]. Different studies have been carried out to determine the ecological function of scorpion fluorescence under UV light, including whether fluorescence can help scorpions to capture prey, attract mates, ward off predators, and identify shelter [7,10,15-16].

In this work, we present a fully automatic and real-time system for the detection of scorpions, based on the implementation of a double validation process using the shape features of the scorpions and the fluorescent characteristics when these arachnids are exposed to UV light. This novel detection method for scorpions was developed using deep learning and computer vision techniques. Deep learning is a research field of machine learning and artificial intelligence that uses multiple processing layers to infer and extract information from big data [17]. In recent years, deep learning techniques have begun to be used in the field of computer vision, in order to acquire, process, analyze, and understand digital images for different applications [18-22].

## 2. Materials and methods

As was mentioned above, a fully automatic and real-time system for the detection of scorpions was developed. In this system, a double detection process was generated from the shape features and the fluorescence property of the scorpions.

For the first detection mechanism, the scorpion shape detection was performed using one of the following two models, known as Haar Cascade Classifier (HCC) method [23] and YOLO (You Only Look Once [24]) method.

On the one hand, the HCC approach is a fast and effective object-detection method based on machine learning technique for image processing. For the training of this model, which has been implemented in OpenCV, a dataset of 2500 images with scorpions (positive images) and 3914 images without scorpions (negative images) has been used in this work. The positive images correspond to many different genera of scorpions, such as *Bothriurus* and *Tityus*, whereas the negative images correspond to other type of arachnid or the absence of specimens. Then, the trained cascade function is used to detect shape features of scorpions in other images.

On the other hand, the YOLO model is based on deep convolutional neural networks [25] for real-time object detection. Compared to other similar models, this model is increasingly used for its high speed and efficiency. In particular, we have used the third version of YOLO (*YOLOv3*) for object detection. For the training of this model, the Keras and TensorFlow libraries have been used [26,27]. Also, the same 2500 positive images were used for this model. However, the previous negative images were not used, since this model generates its own negative images from the parts of the image where the scorpion is not found.

Once the scorpions were detected by their shape characteristics, either by the HCC technique or by the YOLO technique, the detection of the fluorescence emitted by the scorpions under UV light is used as a second detection mechanism in order to confirm their presence. In this case, the study was focused on determining the response of the system to detect the specific fluorescent color of scorpions, to obtain a safer detection system. A deeper analysis of fluorescence detection is presented in the next section.

Fig. 2 shows the flow chart of the system developed in this paper, which consist of a computer, a webcam and a source of UV light. The system can be used at night or in the absence of daylight, therefore the tests were carried out in a dark environment where the fluorescence emitted by the scorpions can be visualized.

In order to validate and better understand the scope of the trained models, confusion matrix were implemented to classify the results and evaluate the performance of these models from the following four metrics: accuracy (A), precision (P), recall (R), and F$_{measure}$ (F1). The following equations are used to calculate these metrics [28]:

$$A = \frac{(TP+TN)}{(TP+TN+FP+FN)} \quad (1)$$

$$P = \frac{TP}{(TP+FP)} \quad (2)$$

$$R = \frac{TP}{(TP+FN)} \quad (3)$$

$$F_{measure} = 2 \cdot \frac{P \cdot R}{(P+R)} \quad (4)$$

where TP, TN, FP and FN denote true positives, true negatives, false positives and false negatives, respectively.

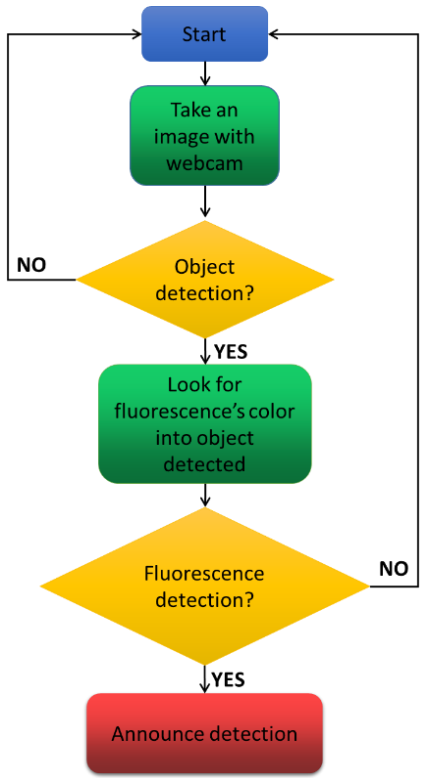

**Figure 2.** Schematic flow chart of the system developed in this work.

The receiver operating characteristic (ROC) curve [29] was also used to examine the behavior of the binary model when the detection threshold is changed. Only the TPs and FPs are necessary to graph the ROC curve, in which the TP rate (sensitivity) is plotted against the FP rate (1-specificity). Every point in a ROC space corresponds to a given instance of the confusion matrix and represents a relative trade-off between TP and FP. The higher the values of TP with respect to FP, the better the trained model will be. The optimal model corresponds to the point located in the upper left corner of the ROC space, with coordinates (0,1). which represents 100% sensitivity (no false negatives) and 100% specificity (no false positives).

## 3. Results and discussion

### 3.1. Analysis of the scorpion shape detection models

First, the performance of both scorpion shape detection models (HCC and YOLO) was analyzed, from images with scorpions (positive images) and without scorpions (negative images) under natural light conditions.

For analysis, a video of 5 frames per second, with a total of 45 seconds, was used. The frames were grouped in 45 blocks of five frames each, therefore one block per second is considered.

Since more than one detection per second no practical sense for the purposes of reaction time in the presence of a scorpion, then, it can be considered that during one second (or one block) the model has correctly detected a scorpion, if at least one frame per block has detected it.

With this criterion, the proportion of the number of seconds with detection, with respect to the total time was calculated, thus obtaining a recall of 73.33% (33 seconds with detection) for the HCC model, and 82.22% (37 seconds with detection) for the YOLO model, which demonstrates its advantage in adverse lighting and camera orientation conditions.

Figures 3 and 4 show an example of detection with the HCC and YOLO models, respectively.

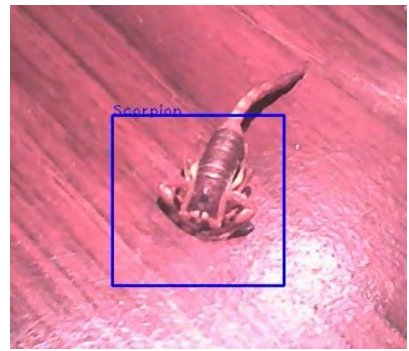

**Figure 3.** Real-time detection of a scorpion using the Haar Cascade Classifier model.

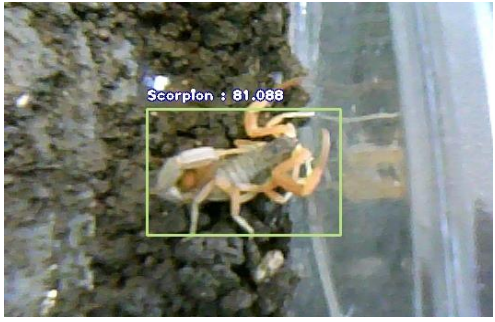

**Figure 4.** Real-time detection of a scorpion using the YOLO model.

### 3.2. Enhancement for fluorescence detection

Fluorescence detection consists of the implementation of digital image morphological processing techniques. In particular, we use the HSV (Hue, Saturation and Value) image format, looking for a range of specific values for the color attribute of each pixel.

To obtain the color of the image, a code was developed, which captures an image from the Webcam and displays the Hue value of the pixel selected with the mouse. Fig. 5 shows a sampling of multiple points of the scorpion with fluorescence, to obtain a range of values to be used as a reference during the detection process.

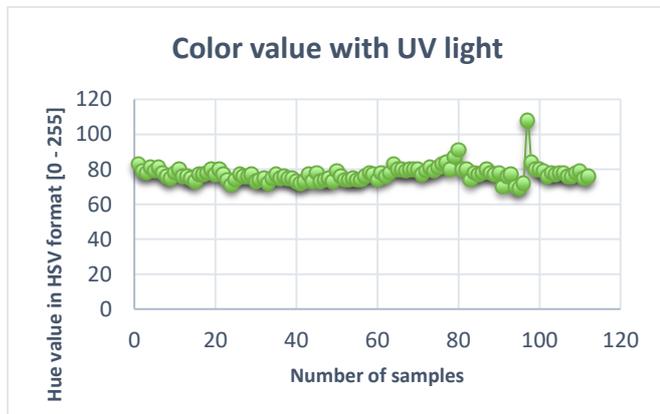

**Figure 5.** Hue values obtained from the color of the scorpion with fluorescence.

An average value of 77.4 and a standard deviation of 4.5 were obtained, therefore, the Hue values range from 73 to 82. This range of values corresponds to the color of the fluorescence emitted by scorpions when they are illuminated with UV light. In each image, once an object is detected, then their coordinates within the image are recorded to be analyzed with the HSV color model. Fig. 6 shows an image of a scorpion considering only the fluorescence color.

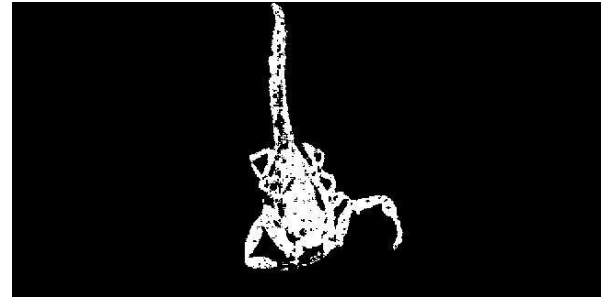

**Figure 6.** Scorpion image only with fluorescence color.

Different tests were performed to study the behavior of the fluorescence detection system within an uncontrolled environment, that is, when the scorpion is close to other objects that could make detection difficult.

Fig. 7 shows an image only with fluorescence color, for the case of a scorpion together with a ball bug (*Porcellio laevis*), which can generally share the same environment. The following problem was found in this case. Although the ball bug does not fluoresce under UV light, it can be seen in this figure the contour of this insect, which is due to the reflection of its wet surface.

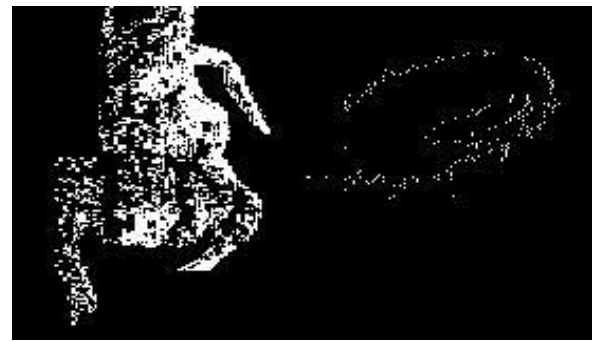

**Figure 7.** Image of a scorpion (left) and a ball bug (right) only with fluorescence color.

In order to eliminate this problem, we use the method of erosion and dilation. Four different combinations are shown in figures 8, 9, 10 and 11, for the cases of a simple erosion, a simple dilation, two dilations and six erosions, and two dilation operations followed by six erosions and eight dilations, respectively.

In the case of a simple initial erosion (figure 8), although the image of the ball bug disappears, the image of the scorpion is also degraded, which could even disappear in those cases with less fluorescence, thus impairing its detection. While in figure 9, with a simple initial dilation, although the brightness of the ball bug is highlighted, the image of the scorpion is strongly consolidated by joining its internal areas.

After two dilations and six erosions (figure 10), the ball bug completely disappeared. In this case, the image of the scorpion was very degraded, but with a sufficient area to recover its original shape and size from new dilations.

The optimal number of operations was obtained for two dilations followed by six erosions and eight dilations (figure 11). After these operations, the ball bug is not observed, while the size and density of the scorpion is suitable for an efficient detection.

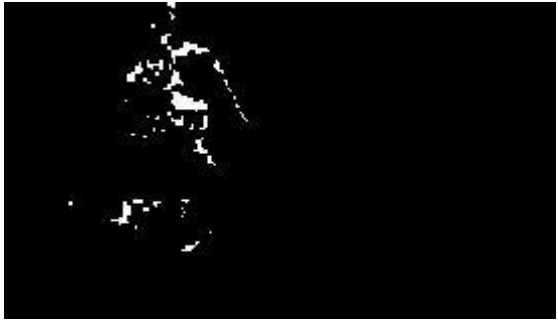

**Figure 8.** Image for a simple erosion operation.

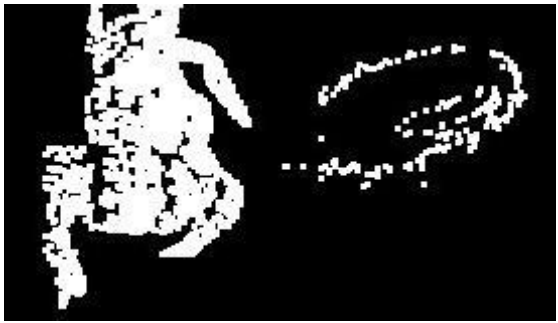

**Figure 9.** Image for a simple dilation operation.

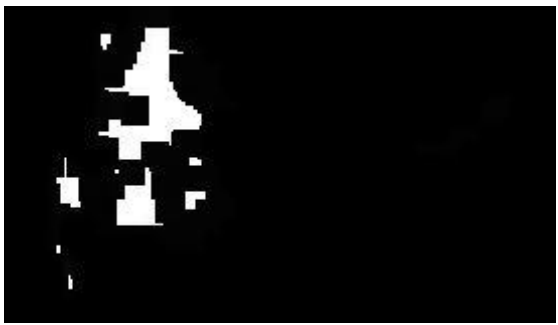

**Figure 10.** Image for two dilations and six erosions.

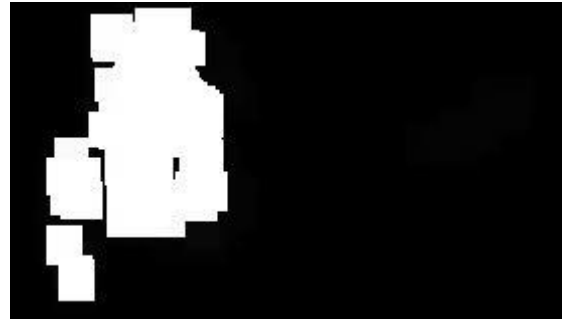

**Figure 11.** Image for two dilations, six erosions and eight dilations.

### 3.3. Analysis of the scorpion detection system with validation process

Finally, the performance of the scorpion detection system, considering the HCC+UV and YOLO+UV methods, was analyzed and compared to each other, from a video both with natural light and in a dark environment.

A true positive (TP) occurs when the fluorescence color is expected, and it is correctly detected. On the contrary, a false positive (FP) occurs when the system wrongly detects the fluorescence color in an image without fluorescence. Whereas a false negative (FN) occurs when the fluorescence color is expected, and the system fails to detect it. Then, the worst situation occurs in the case of FN due to the health risk that this entails.

First, a video of 5 frames per second, with a total of 52 seconds, was used to evaluate the performance of the system when the scorpions are in a dark environment, where the fluorescence emitted by these arachnids can be visualized (that is, possible cases of TP or FN). Since more than one frame per block can correctly detect a scorpion with fluorescence color, then, all frames in which there was detection were considered. These were, 174 and 131 frames (cases of TP) for the HCC+UV and YOLO+UV methods, respectively.

Secondly, the generation of FP was forced, which was achieved using images of scorpions in a natural light environment, situation where it is not possible to detect fluorescence. In order to have a balanced analysis, similar number of frames with detections were used, that is, 174 and 131 FP for the HCC+UV and YOLO+UV methods, respectively. Then, for each successful scorpion shape detection, the fluorescence detection process was analyzed for confirmation.

Starting from initial accuracy and precision of 50%, the use of the second detection mechanism, based on the fluorescent characteristics of the scorpions, allows reducing the FP to only 2 and 5 for the HCC+UV and YOLO+UV methods, respectively.

The confusion matrices obtained during the testing for the HCC+UV and YOLO+UV methods are shown in figures 12

and 13, respectively, where the vertical axes correspond to the true data and the horizontal axes correspond to the predictions of the models.

Table 1 shows the values of accuracy, precision, recall and $F_{measure}$ calculated using equations (1)–(4), respectively, for both methods considered in this study. The high values of the metrics indicate that both methods can successfully detect scorpions. In particular, the values obtained of recall show that there are no false negatives, which is essential for health security purposes.

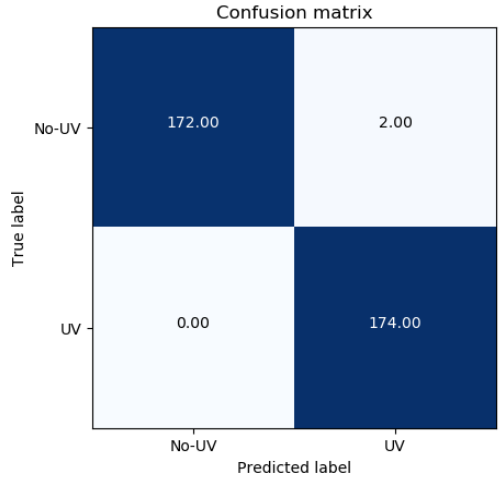

**Figure 12.** Confusion matrix of the HCC+UV method.

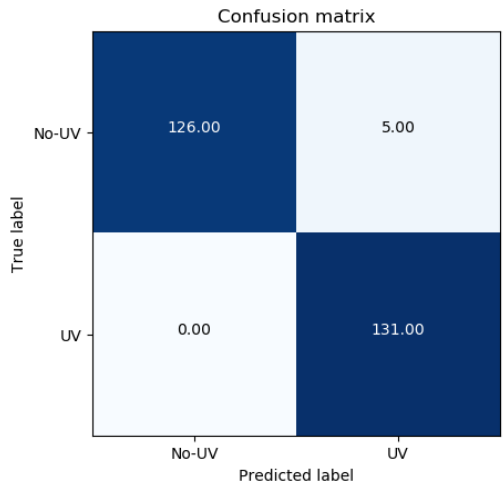

**Figure 13.** Confusion matrix of the YOLO+UV method.

**Table 1.** Metrics for the detection system under study

| Method | Accuracy (*A*) | Precision (*P*) | Recall (*R*) | $F_{measure}$ |
|---|---|---|---|---|
| HCC+UV | 0.99 | 0.99 | 1.00 | 0.99 |
| YOLO+UV | 0.98 | 0.96 | 1.00 | 0.98 |

Fig. 14 shows an example of detection with the YOLO+UV method.

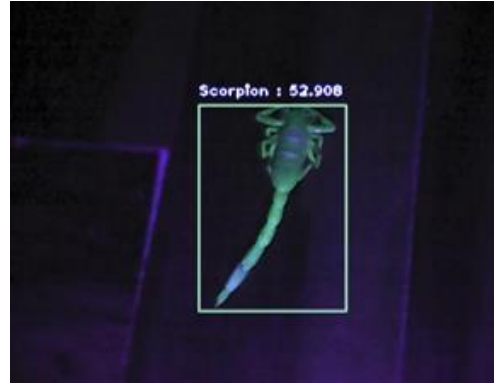

**Figure 14.** Real-time detection of a scorpion using the YOLO+UV method.

Fig. 15 shows the ROC curves for these detection methods. It can be seen in this figure that the areas under the ROC curves for the HCC+UV and YOLO+UV methods are very similar, with area of 99% and 98%, respectively, which implies a very good sensitivity and specificity relationship.

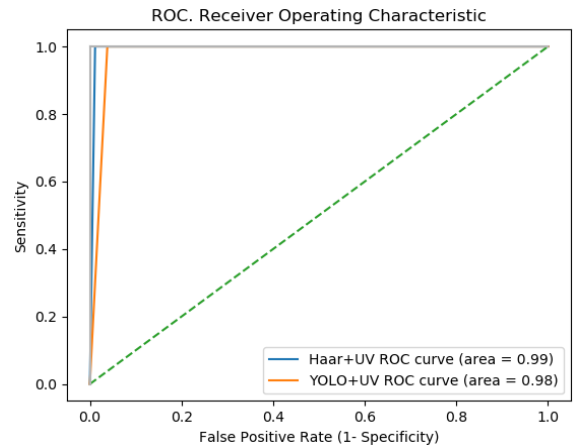

**Figure 15.** Comparative ROC curves for the detection system under study.

## 4. Conclusion

A fully automatic and real-time system capable of detecting scorpions has been proposed in this paper. In this system, a double detection process was generated from the shape features and the fluorescence property of the scorpions when they are exposed to UV light. For the first detection mechanism, based on deep learning and computer vision techniques, the HCC and YOLO models were used and compared for the scorpion shape detection process, which provide recall values of 73.33% and 82.22%, respectively. For the second detection mechanism, HSV image format was used to detect the specific fluorescent color of scorpions (around 500 nm). Different combinations of erosion-dilation were carried out in order to optimize the size and density of the scorpion for an efficient detection within an uncontrolled environment. Finally, the scorpion detection system combining computer vision and fluorescence was analyzed. This system was forced to have an initial accuracy and precision of 50%, which was obtained by generating false positives. The use of the fluorescent property as a second detection mechanism, allows to increase the accuracy and precision to values very close to 100%. The high values of the metrics indicate that both methods (HCC+UV and YOLO+UV) can successfully detect scorpions. In particular, the high values obtained of recall show that the system is an essential tool for health security purposes.